\documentclass[conference]{IEEEtran}
\IEEEoverridecommandlockouts                              
\usepackage{graphicx}
\usepackage{multirow}
\usepackage{amsmath}
\usepackage{booktabs}
\usepackage{adjustbox}
\usepackage{pgfplots}
\usepackage{caption}
\usepackage{subcaption}

\makeatletter
\let\NAT@parse\undefined
\makeatother

\usepackage[breaklinks,colorlinks]{hyperref}

\usepackage[capitalize]{cleveref}
\crefname{section}{Section}{Sections}
\Crefname{section}{Section}{Sections}
\crefname{table}{Table}{Tables
\Crefname{table}{Table}{Tables}}
\crefname{figure}{Fig.}{Figs.}
\Crefname{figure}{Figure}{Figures}

\title{\LARGE \bf
Enhancing Agricultural Environment Perception via \\
Active Vision and Zero-Shot Learning
}

\author{
    \IEEEauthorblockN{Michele Carlo La Greca\textsuperscript{*,1}, Mirko Usuelli\textsuperscript{*,\dag,2}, and Matteo Matteucci\textsuperscript{2}}
    \\
    \IEEEauthorblockA{
        Department of Electronics, Information and Bioengineering\\
        Politecnico di Milano\\
        Milan, Italy\\
        {\tt\footnotesize michelecarlo.lagreca@mail.polimi.it\textsuperscript{1}} \\
        {\tt\footnotesize \{name.surname\}@polimi.it\textsuperscript{2}}
    }
    \thanks{*: Equal Contribution. \dag: Corresponding Author.}
    \thanks{This study was conducted within the Agritech National Research Center and received funding from the European Union Next-GenerationEU (PIANO NAZIONALE DI RIPRESA E RESILIENZA (PNRR) – MISSIONE 4 COMPONENTE 2, INVESTIMENTO 1.4 – D.D. 1032 17/06/2022, CN00000022). This manuscript reflects only the authors’ views and opinions, neither the European Union nor the European Commission can be considered responsible for them.}
}

\begin{document}

\maketitle
\thispagestyle{empty}
\pagestyle{empty}

\begin{abstract}
Agriculture, fundamental for human sustenance, faces unprecedented challenges. The need for efficient, human-cooperative, and sustainable farming methods has never been greater. The core contributions of this work involve leveraging Active Vision (AV) techniques and Zero-Shot Learning (ZSL) to improve the robot's ability to perceive and interact with agricultural environment in the context of fruit harvesting. The AV Pipeline implemented within ROS 2 integrates the Next-Best View (NBV) Planning for 3D environment reconstruction through a dynamic 3D Occupancy Map. Our system allows the robotics arm to dynamically plan and move to the most informative viewpoints and explore the environment, updating the 3D reconstruction using semantic information produced through ZSL models. Simulation and real-world experimental results demonstrate our system's effectiveness in complex visibility conditions, outperforming traditional and static predefined planning methods. ZSL segmentation models employed, such as YOLO World + EfficientViT SAM, exhibit high-speed performance and accurate segmentation, allowing flexibility when dealing with semantic information in unknown agricultural contexts without requiring any fine-tuning process.

\begin{figure}
    \centering
    \includegraphics[scale=0.27]{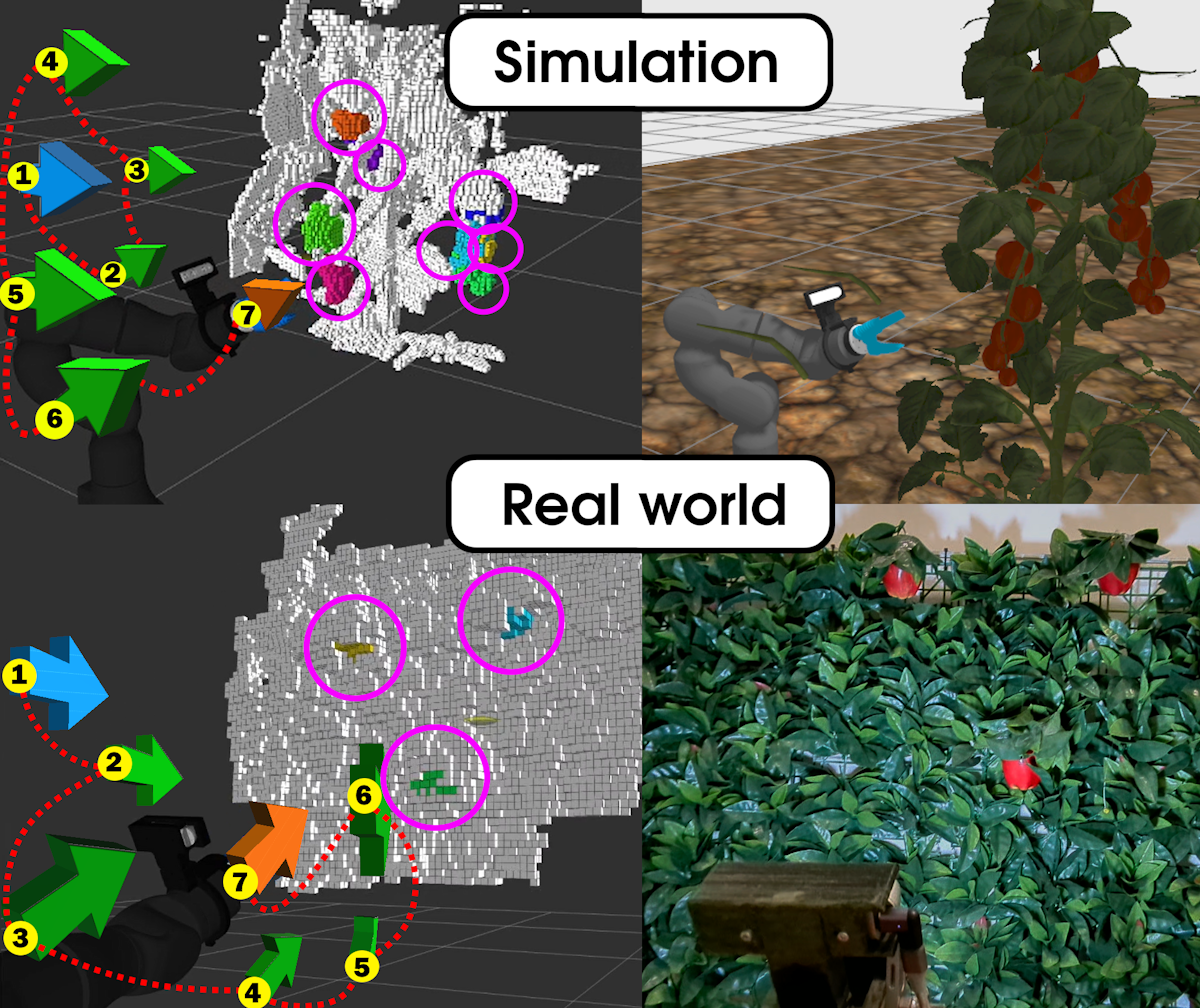}
    \caption{3D reconstruction using Active Vision. The upper figure depicts the process for a simulated tomato plant, while the lower figure shows the NBV planning for a real espalier apple tree with occluded fruits.}
    \label{fig:pointcloudpoints}
    \vspace{-1.5em}
\end{figure}

\end{abstract}


\section{Introduction}

Agriculture is essential to society as it is responsible for the primary source of food that sustains human life. With the demand for food increasing, agriculture must evolve to produce higher yields while maintaining the sustainability of natural resources such as land and water~\cite{VIANA2022150718}. Robotics, capable of performing repetitive and labor-intensive tasks such as planting, harvesting, and weeding, can boost productivity in agriculture, minimizing waste, maximizing crop yields, improving working conditions, and thus reducing the physical strain on human laborers. To successfully integrate robotics into agriculture and enhance their performance, the initial priority is enabling robots to accurately perceive the unstructured rural environment. The agricultural environment is frequently complex and dynamic, with plants and crops that may be obscured or have intricate, varied structures. This complexity presents substantial challenges for traditional management methods, underscoring the need for advanced perception approaches to improve the deployability of automation.

The proposed research introduces an approach to overcome the challenges and complexities of fruit perception through Active Vision (AV). Instead of relying on passive observation, AV allows robots to actively perceive, explore, and reconstruct at run-time their surroundings by planning the optimal position of the camera viewpoint using the Next-Best View (NBV) planning~\cite{pito1999solution} which maximizes the information gained regarding plants and crops. This ensures that even hidden or occluded parts of the environment are effectively captured (Fig.~\ref{fig:pointcloudpoints}). To achieve this, semantic information about the plants and crops needs to be exploited for informative guidance. In this respect, this work applies Zero-Shot Learning (ZSL)~\cite{xian2018zero} to provide useful segmentation, enabling the robot to generalize and adapt to various crops or environmental features without requiring specific training data for each scenario. By leveraging both 3D and semantic data, the robot can reconstruct a detailed, semantic, and context-aware map of the environment. This enhanced understanding allows the robot to strategically adjust its movements and positioning, leading to more effective interactions with the environment.

This research focuses on the following contributions:
\begin{itemize}
    \item Developed a modular architecture in C++ and ROS 2 for Active Vision in agricultural robotics, addressing the challenge of detecting occluded fruits during harvest;
    \item To the best of the author’s knowledge, this is the first work to integrate zero-shot learning perception with Active Vision exploration, enabling environment-independent operation in agriculture;
    \item Conducted extensive evaluations both in simulation and real-world scenarios, in contrast to state-of-the-art methods that primarily focus on simulated environment with supervised learning.
    \item Set a benchmark standard for the lack of reproducibility and availability of open source code in the context of Active Vision in agricultural robotics.
\end{itemize}
The entire implementation of this work is open source and available on GitHub at ~\url{https://github.com/AIRLab-POLIMI/active-vision}.

\vspace{-1.0em}
\section{Related works}
Active Vision enables robots to optimize their visual sensors' positions to gather useful information for various tasks~\cite{Zeng2020}. Traditional approaches often struggle with efficiently identifying relevant objects in complex settings, particularly in agricultural applications like greenhouse environment, where occlusion is a significant challenge~\cite{burusa2024semanticsawarenextbestviewplanningefficient}. 

Burusa et al.~\cite{burusa2024semanticsawarenextbestviewplanningefficient} have recently aimed to improve the efficiency of AV systems by incorporating semantic information into viewpoint planning. They introduced semantics-aware strategies that prioritize task-relevant plant components, such as tomatoes, peduncles, and petioles, during the view planning process. Their approach features a modular pipeline consisting of sensing modules detecting objects of interest (OOIs) using Mask R-CNN~\cite{DBLP:journals/corr/HeGDG17} specifically fine-tuned on tomatoes in simulation. It also includes 3D scene representation modules with attention mechanisms for tracking these OOIs across multiple viewpoints, and view-planning modules that determine the next-best viewpoint to optimize task performance. Additionally, in their work, clustering is applied for instance refinement at the voxel-map level. 

Kriegel et al.~\cite{6696691} introduce a system that integrates AV by analyzing scenes with depth images, identifying and segmenting objects which are then stored in a dynamic database. In their work, iterative view planning improves object recognition and modeling by acquiring multiple views to reduce occlusions and enhance accuracy. A probabilistic voxel space is used to represent explored and unexplored regions, aiding in collision-free path planning, selecting minimal occlusion views, and verifying object poses. The system also autonomously models unknown objects, adding them to the database for future recognition.

AV has been significantly advanced through the integration of Reinforcement Learning (RL) techniques~\cite{10.5555/3312046}, where models are trained to map situations to actions, optimizing decisions to maximize a numerical reward. Ammirato et al.~\cite{ammirato2017datasetdevelopingbenchmarkingactive} have introduced extensive datasets of images and object detection bounding boxes captured from diverse real-world scenes, providing simulations of robot interactions in various environments. These datasets are fundamental for training and evaluating deep neural networks aimed at predicting optimal moves to enhance object recognition accuracy through RL. Moreover, AV has been further refined using Recurrent Neural Networks (RNNs)~\cite{10.5555/525960}, which excel at modeling temporal dependencies in sequential visual data. Xu et al.~\cite{10.1145/2980179.2980224} demonstrated autonomous object exploration and identification in indoor environment using 3D attention models. Their approach leverages deep recurrent networks to handle temporal sequences and hierarchical classifiers to accurately identify objects from a large 3D shape collection, prioritizing the most informative views and regions.

Traditional methods such as SVMs, decision trees, K-means, and CRFs were widely used for image segmentation tasks before the rise of Deep learning~\cite{fi15060205}. Modern approaches focus on semantic and instance segmentation~\cite{10.1007/978-981-97-4387-2_11}, with the former assigning category IDs to pixels and the latter identifying individual objects within the same category. Recent advancements, such as Zero-Shot Learning (ZSL)~\cite{10.1145/3293318}, allow models to recognize unseen classes using auxiliary semantic information, such as text, reducing the need for extensive labeled supervised datasets. Notable innovations such as Segment Anything (SAM)~\cite{sam}, Efficient SAM~\cite{xiong2023efficientsamleveragedmaskedimage}, Grounding DINO~\cite{liu2024groundingdinomarryingdino}, and YOLO World~\cite{yoloworldrealtimeopenvocabularyobject} push the boundaries by generalizing across new image distributions and tasks without requiring task-specific training.

Building on the advancements in segmentation, these improvements also influence 3D reconstruction techniques, which rely heavily on accurate object identification. Probabilistic 3D maps, such as OctoMap~\cite{octomap}, are fundamental for fast 3D reconstruction in robotics. These maps use octrees to efficiently represent and update environment, optimizing memory and computation. Recent methods such as BONXAI~\cite{bonxai} further reduce memory usage without predefined boundaries, dynamically adjusting to data density~\cite{bonx}. Regardless of the data structure, semantic information can be associated with each 3D block and used in NBV strategies, as demonstrated in the literature~\cite{burusa2024semanticsawarenextbestviewplanningefficient, 10.1145/2980179.2980224}.

\vspace{-0.1em}
\section{The Proposed Approach}
\begin{figure*}[ht]
    \vspace{1.5em}
    \centering
    \includegraphics[width=0.95\textwidth]{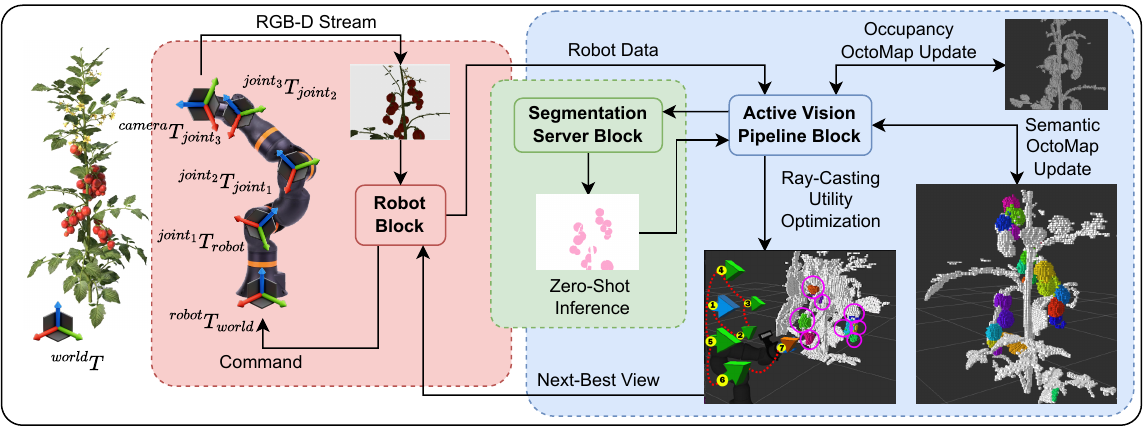}
    \caption{The diagram shows a 6-DoF robotic arm with a camera mounted on top. The system explores the agricultural environment by retrieving state data from the Robot Block to the Active Vision Pipeline Block. This pipeline performs ZSL segmentation via the Segmentation Server Block and updates the Semantic 3D Occupancy Map. Following the update, the ray-casting utility optimization is activated to generate the NBV, which is used for the final loop closing control of the robot.}
    \label{fig:overview}
    \vspace{-0.9em}
\end{figure*}

\vspace{-0.1em}
The proposed approach features a modular architecture
organized into three main components: (1) Robot Block, (2) Segmentation Server Block, and (3) Active Vision Pipeline Block, as shown in Fig.~\ref{fig:overview}. Each component integrates various functionalities to obtain a responsive robotics system, enabling AV for advanced environmental perception.

\subsection{Overall Architecture}

The proposed AV system is built as a multi-threaded and distributed architecture, allowing the deployment of its components across multiple machines, including:
\begin{itemize}
    \item The Robot Block manipulates and communicates the robot’s current physical state to other nodes. It also integrates motion planning components, which work together to plan and perform robot movements while interfacing with low-level hardware components such as joints, sensors, and actuators. Lastly, it provides the sensor data necessary for environmental perception.
    \item The Segmentation Server Block is responsible for handling all segmentation requests from the client, i.e. the Active Vision Pipeline Block. It operates the ZSL models to perform inference on-demand. Thanks to its server capabilities, the system can communicate and provide results without wasting resources during run-time. Being independent, it is easily deployable on a dedicated GPU and can seamlessly communicate with other machines.
    \item The Active Vision Pipeline Block manages perception by integrating data from segmented images, requested from the Segmentation Server Block, point clouds extracted from the depth channel of the captured image, and the stored semantic 3D occupancy map, which is updated within the block's logic. Based on the 3D semantic reconstruction, the NBV is determined to guide the robot operating the MoveIt2 API~\cite{Giampa2024} for motion planning. This approach minimizes latency and enhances performance, particularly with dense point clouds, thanks to multi-threaded operations performed.
\end{itemize}

\vspace{-0.7em}
\subsection{Active Vision Pipeline}
\vspace{-0.1em}

The AV pipeline consists of executing a loop as illustrated in Fig.~\ref{fig:overview}, where perception and planning alternate, updating the semantic 3D occupancy map that is used for NBV planning.

\subsubsection{Segmentation as a Service}

Segmentation is conducted using a client-server paradigm to alleviate bottlenecks in the image stream, allowing the system to focus on discretized viewpoints as the robot moves. Upon receiving sensor data, the client sends a request containing an image, a text prompt to guide the segmentation process, and confidence thresholds. The server processes this request, performs the inference, and returns the segmented results.

Two ZSL segmentation approaches are integrated as servers in this work: 

\begin{itemize}
    \item Lang SAM (\textbf{LSAM})~\cite{medeiros2024langsegmentanything}: it integrates two models for performing ZSL segmentation: Grounding DINO and SAM. Grounding DINO is a model that detects objects in an image based on a text prompt. The bounding box produced is then passed to SAM, which creates the segmented masks and related confidences of the unknown classes.
    
    \item YOLO World + EfficentViT SAM (\textbf{YWES})~\cite{park2024yoloworld}: The same concept applies to YOLO World + EfficientViT SAM, a combination of YOLO World, an open-vocabulary object detection model, and EfficientViT SAM, a new family of accelerated SAM models.
\end{itemize}

Upon receiving the client request, the server performs inference by feeding the detection bounding boxes generated by the detection model into the segmentation model. This inference produces a mask and confidence score for each detected instance. The results, along with their corresponding semantic classes, are then packaged into a custom message and sent back to the client speeding up the system workflow.

\begin{table*}[th!]
\vspace{1.5em}
  \centering
  \begin{tabular}{@{}ccccccccccc@{}}
  \toprule
    \textbf{Prompt} & \textbf{Model} & \textbf{Inference Time (\textit{s})} & \textbf{Detection Conf. (\%)} & \textbf{Accuracy (\%)} & \textbf{Precision (\%)} & \textbf{Recall (\%)} & \textbf{F1 Score (\%)} \\
    \midrule
    \texttt{`apple`} & LSAM & 18.8755 & 69.48 & 100.00 & 100.00 & 100.00 & 100.00 \\
      & YWES & 4.1357 & 95.03 & 100.00 & 100.00 & 100.00 & 100.00  \\
    \midrule
    \texttt{`green apple`} & LSAM & 18.6254 & 35.76 & 100.00 & 100.00 & 100.00 & 100.00 \\
      & YWES & 2.6751 & 89.99 & 50.00 & 50.00 & 100.00 & 66.67 \\
    \midrule
    \texttt{`tomato`} & LSAM & 17.6048 & 65.36 & 100.00 & 100.00 & 100.00 & 100.00 \\
      & YWES & 3.7633 & 18.03 & 100.00 & 100.00 & 100.00 & 100.00 \\
    \midrule
    \texttt{`mature tomato`} & LSAM & 15.6835 & 62.57 & 40.00 & 40.00 & 100.00 & 57.14 \\
      & YWES & 2.9241 & 12.80 & 40.00 & 40.00 & 100.00 & 57.14 \\
    \midrule
    \texttt{`berry`} & LSAM & 13.8146 & 39.49 & 83.00 & 83.00 & 100.00 & 90.91 \\
      & YWES & 3.0246 & 6.49 & 100.00 & 100.00 & 100.00 & 100.00 \\
    \bottomrule
  \end{tabular}
  \caption{Performance comparison of ZSL models in various experiments focused on stand-alone segmentation.}
  
  \label{tab:standaloneinference}
  \vspace{-0.9em}
\end{table*}

\subsubsection{3D Occupancy Map Creation}

After receiving the segmentation response, point clouds are generated from the sensor's RGB image and depth data, which are used to update the occupancy information. Semantic instances are added to the 3D occupancy map by projecting the ZSL masks onto the depth image, including confidence scores and semantic classes.

The 3D occupancy map is created and updated by first defining the Octree structure and integrating the sensor's point cloud to update or add voxels based on occupancy. Unoccupied voxels are excluded from calculations to reduce computational load, though this may lead to inconsistencies in dynamic environments where unoccupied space needs updating.

For semantic integration, a hash map is employed where voxel addresses serve as keys, with values containing semantic class, confidence, instance number, and additional details. Point clouds from segmented masks are used to update the semantic data of each voxel. The voxel’s semantic information is initially set, and for each segmented point cloud, the semantic class and confidence are updated using the max fusion algorithm~\cite{burusa2024semanticsawarenextbestviewplanningefficient}. Outlier detection further enhances the quality of the 3D reconstruction.

\begin{figure}[ht!]

\begin{subfigure}{0.48\textwidth}  
    \begin{minipage}{0.55\textwidth}  
        \begin{tikzpicture}
            \begin{axis}[
                title=\textbf{Full Occlusion},
                title style={yshift=-7pt},
                width=\textwidth,
                height=\textwidth,
                xlabel={Planning steps},
                ylabel={F1 Score},
                xlabel style={yshift=5},
                ylabel style={yshift=-10pt},
                xmin=1, xmax=8,
                ymin=0, ymax=0.4,
                xtick={1,2,3,4,5,6,7,8},
                ytick={0,0.1,0.2,0.3,0.4},
                yticklabel style={/pgf/number format/fixed, /pgf/number format/precision=2},
                legend pos=north west,
                legend style={font=\scriptsize},
                ymajorgrids=true,
                grid style=dashed,
            ]

            \addplot[
                color=blue,
                mark=square,
                ]
                coordinates {
                (1,0.083914)(2,0.083914)(3,0.083914)(4,0.083914)(5,0.142127)(6,0.142127)(7,0.148513)(8,0.148513)
                };
            \addlegendentry{Predefined}

            \addplot[
                color=red,
                mark=triangle,
                ]
                coordinates {
                (1,0.083610)(2,0.100033)(3,0.114767)(4,0.125926)(5,0.128036)(6,0.219203)(7,0.285220)(8,0.317657)
                };
            \addlegendentry{Our approach}

            \end{axis}
        \end{tikzpicture}
    \end{minipage}
    \begin{minipage}{0.4\textwidth}  
        \centering
        \includegraphics[width=\textwidth]{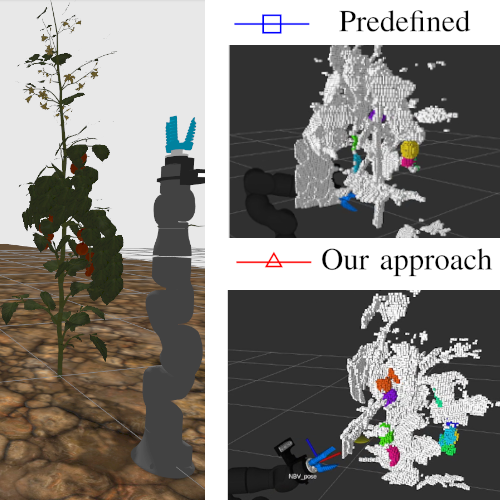}
    \end{minipage}
    \vspace{-0.25em}
    \caption{Vegetation that fully obscures the tomato fruits, testing its ability to detect and explore hidden clusters.}
    \label{fig:scene1}
\end{subfigure}
\vfill
\begin{subfigure}{0.48\textwidth}  
    \begin{minipage}{0.55\textwidth}  
        \begin{tikzpicture}
        \begin{axis}[
            title=\textbf{Multiple Grapes},
            title style={yshift=-7pt},
            width=\textwidth,
            height=\textwidth,
            xlabel={Planning steps},
            ylabel={F1 Score},
            xlabel style={yshift=5},
            ylabel style={yshift=-10pt},
            xmin=1, xmax=8,
            ymin=0.4, ymax=0.8,
            xtick={1,2,3,4,5,6,7,8},
            ytick={0.4, 0.5, 0.6, 0.7, 0.8},
            yticklabel style={/pgf/number format/fixed, /pgf/number format/precision=2},
            legend pos=north west,
            legend style={font=\scriptsize},
            ymajorgrids=true,
            grid style=dashed,
        ]
        \addplot[
            color=blue,
            mark=square,
            ]
            coordinates {
            (1,0.422543)(2,0.570048)(3,0.612173)(4,0.622135)(5,0.641360)(6,0.642140)(7,0.644689)(8,0.644850)
            };
        \addlegendentry{Predefined}

        \addplot[
            color=red,
            mark=triangle,
            ]
            coordinates {
            (1,0.459853)(2,0.577522)(3,0.629450)(4,0.640568)(5,0.651345)(6,0.653727)(7,0.664680)(8,0.678433)
            };
        \addlegendentry{Our approach}

        \end{axis}
    \end{tikzpicture}
    \end{minipage}
    \begin{minipage}{0.4\textwidth}  
        \centering
        \includegraphics[width=\textwidth]{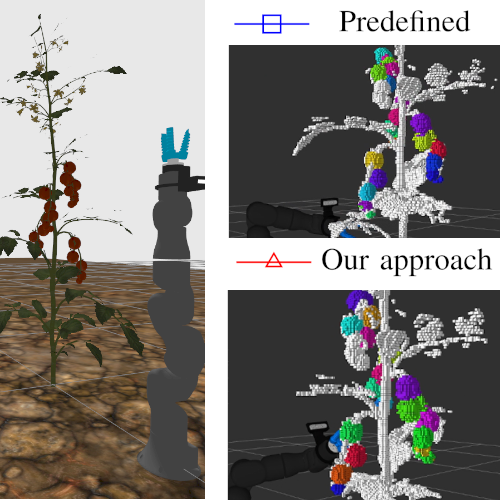}
    \end{minipage}
    \vspace{-0.25em}
    \caption{Evenly distributed tomato grapes, assessing the system’s capacity to explore and reconstruct multiple targets.}
    \label{fig:scene2}
\end{subfigure}
\vfill
\begin{subfigure}{0.48\textwidth}  
    \begin{minipage}{0.55\textwidth}  
        \begin{tikzpicture}
            \begin{axis}[
                title=\textbf{Single Grape},
                title style={yshift=-7pt}, 
                width=\textwidth,
                height=\textwidth, 
                xlabel={Planning steps},
                ylabel={F1 Score},
                xlabel style={yshift=5}, 
                ylabel style={yshift=-10pt}, 
                xmin=1, xmax=8,
                ymin=-0.1, ymax=0.9,
                xtick={1,2,3,4,5,6,7,8},
                ytick={-0.1, 0.1, 0.3, 0.5, 0.7},
                yticklabel style={/pgf/number format/fixed, /pgf/number format/precision=2},
                legend pos=north west,
                legend style={font=\scriptsize},
                ymajorgrids=true,
                grid style=dashed,
            ]

            \addplot[
                color=blue,
                mark=square,
                ]
                coordinates {
                (1,0.000000)(2,0.469565)(3,0.516077)(4,0.516077)(5,0.516077)(6,0.516077)(7,0.516077)(8,0.516077)
                };
            \addlegendentry{Predefined}

            \addplot[
                color=red,
                mark=triangle,
                ]
                coordinates {
                (1,0.000000)(2,0.000000)(3,0.000000)(4,0.000000)(5,0.376623)(6,0.415502)(7,0.531635)(8,0.546995)
                };
            \addlegendentry{Our approach}
            \end{axis}
        \end{tikzpicture}
    \end{minipage}
    \begin{minipage}{0.4\textwidth}  
        \centering
        \includegraphics[width=\textwidth]{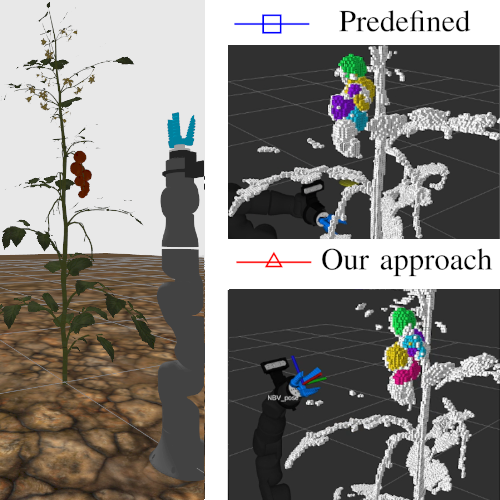}
    \end{minipage}
    \vspace{-0.25em}
    \caption{Naked tomato plant with only one visible grape cluster, focusing on its precision in minimal target environment.}
    \label{fig:scene3}
\end{subfigure}
\vfill
\begin{subfigure}{0.48\textwidth}  
    \begin{minipage}{0.55\textwidth}  
        \begin{tikzpicture}
            \begin{axis}[
                title=\textbf{Unoriented Start},
                title style={yshift=-7pt}, 
                width=\textwidth,
                height=\textwidth, 
                xlabel={Planning steps},
                ylabel={F1 Score},
                xlabel style={yshift=5}, 
                ylabel style={yshift=-10pt}, 
                xmin=1, xmax=8,
                ymin=0.0, ymax=0.9,
                xtick={1,2,3,4,5,6,7,8},
                ytick={0.0, 0.2, 0.4, 0.6, 0.8},
                yticklabel style={/pgf/number format/fixed, /pgf/number format/precision=2},
                legend pos=north west,
                legend style={font=\scriptsize},
                ymajorgrids=true,
                grid style=dashed,
            ]

            \addplot[
                color=blue,
                mark=square,
                ]
                coordinates {
                (1,0.057821)(2,0.057821)(3,0.316957)(4,0.316957)(5,0.361738)(6,0.361738)(7,0.378786)(8,0.378786)
                };
            \addlegendentry{Predefined}

            \addplot[
                color=red,
                mark=triangle,
                ]
                coordinates {
                (1,0.040383)(2,0.449415)(3,0.515541)(4,0.560658)(5,0.580070)(6,0.582978)(7,0.585673)(8,0.588921)
                };
            \addlegendentry{Our approach}

            \end{axis}
        \end{tikzpicture}
    \end{minipage}
    \begin{minipage}{0.4\textwidth}  
        \centering
        \includegraphics[width=\textwidth]{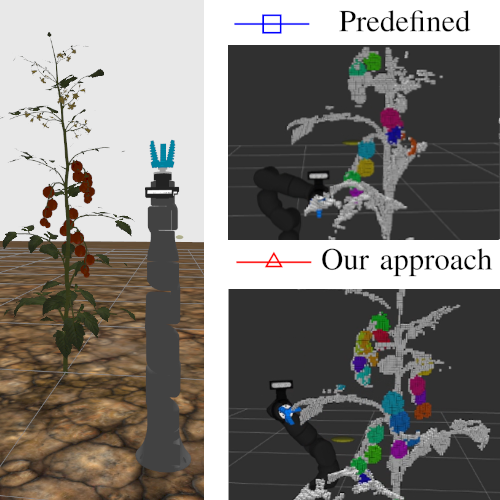}
    \end{minipage}
    \vspace{-0.25em}
    \caption{Misaligned pose relative to the plant, complicating the planner's task of adjusting exploration capabilities.}
    \label{fig:scene4}
\end{subfigure}
\caption{Qualitative comparison of exploration planning convergence across four scenarios of a simulated tomato plant.}
\end{figure}

\subsubsection{Planning}
The final step of the AV pipeline is selecting the NBV after the perception phase of each iteration. The process begins by sampling candidate viewpoints and evaluating them to find the most informative one. For each candidate, a camera frustum is generated to define the visible space. An adjustable attention region focuses the evaluation on specific semantic parts of the 3D occupancy map. Ray casting determines which voxels in the attention region are visible and occupied. The efficiency of different ray casting methods is assessed, with a focus on semantically important areas. The utility of each viewpoint is then calculated based on the \textit{Expected Semantic Information Gain} ($G_{\text{sem}}(\xi)$), computed as the sum of the semantic information of all voxels visible from viewpoint \( \xi \), as in Equation~\ref{eq:sem_inf_gain}:

\begin{equation}
G_{\text{sem}}(\xi) = \sum_{x \in (X_{\xi} \cap B)} I_{\text{sem}}(x),
\label{eq:sem_inf_gain}
\end{equation}
        where \( X_{\xi} \) represents the set of voxels visible from viewpoint \( \xi \), and \( B \) denotes the voxels within the regions of interest. The expected semantic information \( I_{\text{sem}}(x) \) for a voxel \( x \) is defined by its entropy, as shown in Equation~\ref{eq:entropy}:

\begin{equation}
\begin{split}
I_{\text{sem}}(x) = -p_s(x) \log_2(p_s(x)) +\\
- (1 - p_s(x)) \log_2(1 - p_s(x)),
\label{eq:entropy}
\end{split}
\end{equation}
where \( p_s(x) \) is the confidence score for voxel \( x \).

\textit{Expected Semantic Information Gain} reflects the amount of new knowledge regarding the environment that would be obtained by observing a specific voxel. Once the utility of the current pose is computed, the system checks if this pose represents the best viewpoint identified so far. Finally, the viewpoint with the highest utility is selected for the next pose, moving the robot accordingly.

\vspace{-0.7em}
\section{Evaluation}
\vspace{-0.5em}
We conducted experiments to evaluate our system in both real-world and simulated environments. ZSL segmentation with LSAM and YWES was tested independently and within the ROS 2 framework. Additionally, AV for 3D reconstruction in agricultural settings was evaluated using predefined zig-zag movements and our proposed NBV planning approach.

\vspace{-0.5em}
\subsection{Experimental Setup}

\paragraph{\textbf{Robot Platform}} The work utilizes the Igus ReBeL 6-DoF robotic arm~\cite{igusrebel}, both in real-world and simulation settings. This lightweight, cost-effective arm is designed for collaborative tasks and offers flexibility through open-source control, though it has limitations in reach and precision due to its plastic construction.

\paragraph{\textbf{Real-world Implementation}} A hardware interface~\cite{Giampa2024} converts ROS 2 commands into signals for the robotic arm, while a Realsense D435~\cite{d435} camera provides environmental perception. Testing is conducted with an espalier apple tree setup, which is efficient for large-scale production due to its space-saving and yield-enhancing benefits in the primary sector.

\paragraph{\textbf{Simulation Setup}} The simulation replicates the real-world configuration in Gazebo Ignition, using a hardware interface~\cite{gz_ros2_control_github}~\cite{ign_moveit2_examples_github} and virtual sensors that mimic the Realsense D435. It features virtual tomato plants at different vegetation stages to test the approach under controlled conditions similar to the physical setup.

Experiments were conducted on a Dell XPS 13 9305 with 16 GiB of memory, an Intel i7-1165G7 processor, Intel Xe Graphics, and Ubuntu 22.04.4 LTS. The workstation also had ROS 2 Humble and Gazebo Ignition Fortress for simulations.

\begin{table*}[ht]
\vspace{1.5em}
\centering
\begin{minipage}{0.35\textwidth}
\centering
\hspace{-0.15\textwidth}
\includegraphics[width=0.75\textwidth]{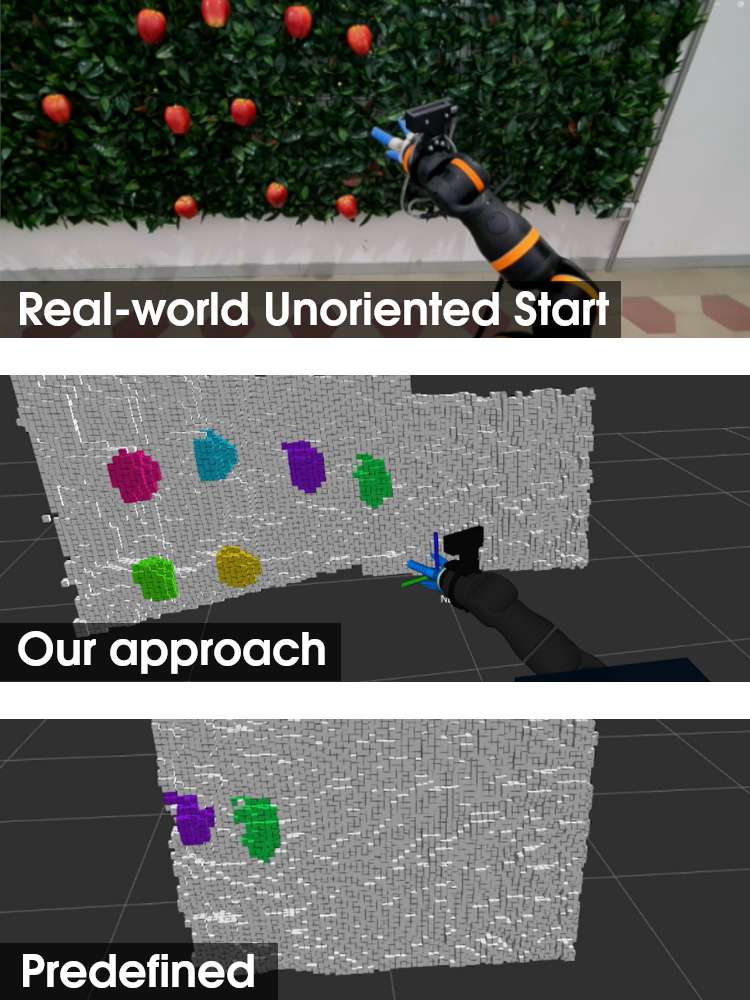} 
\captionof{figure}{Real-world visual comparison \\regarding unoriented start conditions.}
\label{fig:experimental_setup}
\end{minipage}
\begin{minipage}{0.6\textwidth}
\centering

\begin{tabular}{@{\hskip 10pt}ccccc@{\hskip 3pt}}
\toprule
\textbf{Experiment} & \textbf{Environment} & \textbf{Planning} & \textbf{Fruits (\#)} & \textbf{F1 Score (\%)} \\
\midrule
Full Occlusion & Simulation & Predefined & 6/24 & 14.2 \\
               &            & \textbf{Our approach} & \textbf{9/24} & \textbf{31.8} \\
               & Real-world & Predefined & 1/3 & 25.6 \\
               &            & \textbf{Our approach} & \textbf{3/3} & \textbf{43.6} \\
\midrule
Multiple Grapes/Fruits & Simulation & Predefined & 18/24 & 66.7 \\
                       &            & \textbf{Our approach} & \textbf{21/24} & \textbf{67.8} \\
                       & Real-world & Predefined & 5/11 & 49.6 \\
                       &            & \textbf{Our approach} & \textbf{6/11} & \textbf{52.9} \\
\midrule
Single Grape & Simulation & Predefined & 4/8 & 51.6 \\
             &            & \textbf{Our approach} & \textbf{5/8} & \textbf{54.7} \\
\midrule
Unoriented Start & Simulation & Predefined & 10/24 & 37.9 \\
                 &            & \textbf{Our approach} & \textbf{17/24} & \textbf{58.9} \\
                 & Real-world & Predefined & 2/10 & 21.3 \\
                 &            & \textbf{Our approach} & \textbf{6/10} & \textbf{63.8} \\
\bottomrule
\end{tabular}
\caption{Quantitative results from simulations involving tomato plants with grapes and real-world experiments with apple espalier featuring single fruits, analyzed across diverse agronomical scenarios.}
\label{tab:merged_quantitative_results}
\end{minipage}
\vspace{-0.9em}
\end{table*}

\subsection{Experiments}
It is important to note that a direct comparison with the state-of-the-art has not been feasible. This limitation arises primarily from the unavailability of open-source code provided by authors in the relevant literature. Additionally, the high complexity and specialized nature of these systems made it impractical to replicate each relevant system within the domain of agricultural robotics. Consequently, the experimental analysis is focused on evaluating the proposed system's components, specifically in terms of ZSL perception and NBV planning through ray-casting utility optimization across sim-to-real scenarios.

The initial experiments involved ZSL segmentation using LSAM and YWES models conducted outside our framework. These experiments focused on performing inference on a variety of fruit categories, including \texttt{`apple`}, \texttt{`green apple`}, \texttt{`tomato`}, \texttt{`mature tomato`}, and \texttt{`berry`}. Each model was evaluated using standard classification metrics: accuracy, precision, recall, and F1-score. Based on the results, the YWES model, which achieved the best overall performance, was selected for further deployment. Subsequent experiments integrated ZSL within our system during the 3D occupancy map creation process. This involved AV, where the robot captured multiple sensor images for segmentation across different viewpoints:

\paragraph{\textbf{Simulation experiments}} We tested the AV system on tomato plants characterized by multiple clusters of tomatoes. Four distinct scenarios were defined: (1) \textit{Full Occlusion} (Fig.~\ref{fig:scene1}), where vegetation fully obscured the fruits, (2) \textit{Multiple Grapes} (Fig.~\ref{fig:scene2}), where the tomato plant displayed an even distribution of grape clusters, (3) \textit{Single Grape} (Fig.~\ref{fig:scene3}), where only a single cluster of tomatoes was present on a spoiled plant, and (4) \textit{Unoriented Start} (Fig.~\ref{fig:scene4}), in which the robot arm began from a pose misaligned with the plant, thereby stressing the NBV planning algorithm's ability to optimize exploration and discover the grapes.

\paragraph{\textbf{Real-world experiments}} We constructed an apple espalier setup in our laboratory with dense vegetation. Three scenarios were tested: (1) \textit{Full Occlusion}, where apples were minimally visible even to the human eye, (2) \textit{Multiple Fruits}, with a variety of apples distributed across the espalier wall, and (3) \textit{Unoriented Start}, to ensure consistency with the simulation outcomes and validate the convergence capability of the NBV planning algorithm.

In both real-world and simulated environments, the AV system conducted 14 experiments for 3D reconstruction using predefined and NBV planning. Predefined planning involved eight zig-zag poses, while the NBV planning algorithm used a similar number of poses, except for the full occlusion real-world scenario with six poses. The AV pipeline's accuracy was assessed with an F1-score, comparing the ground truth to the reconstructed voxels in the semantic space.

\vspace{-0.9em}
\subsection{Quantitative Results}
Table~\ref{tab:standaloneinference} presents the performance of ZSL segmentation models in identifying various fruits using CPU. YWES significantly outperforms the others with faster inference times, while LSAM is about 4.5 times slower. The classification metrics demonstrate YWES's robustness in object identification across different forms, leading to its selection for deployment on the proposed system.

Table~\ref{tab:merged_quantitative_results} breaks down the quantitative results from simulated and real-world experiments. The number of fruits is evaluated through human visual inspection, showing that our system outperforms the handcrafted baseline in all scenarios. The best results are observed in the \textit{Full Occlusion} and \textit{Unoriented Start} scenarios. While \textit{Full Occlusion} was the primary goal of this research, \textit{Unoriented Start} further validates the need for informative guidance in recovery behaviors for agricultural tasks. Both multiple and single fruit scenarios showed similar scores in F1 Score and fruit counting, with our approach consistently surpassing the baseline. The ZSL and modular pipeline ensure consistent performance across different crops and operating environments.

The fruit count in some experiments was constrained by the fixed range of the robotic arm, despite using human-based counting for metrics assessment. Future work will focus on deploying the system on an agricultural robot to enhance coverage, control the mobile platform with NBV planning, and address the limitations of the robotic arm's operational space identified in this benchmark. Although all experiments were conducted on the CPU, as denoted by inference times in Table~\ref{tab:standaloneinference}, GPU acceleration has the potential to significantly improve inference in future research for real-time performance on a mobile platform.

\vspace{-0.3em}
\subsection{Qualitative results}
\begin{enumerate}
    \item \textit{Full Occlusion} (Fig.~\ref{fig:scene1}): In the majority of planning steps, the NBV planning approach outperforms the baseline in informative exploration. The semantic 3D occupancy map is notably richer due to the extensive vegetation, as highlighted by the quantitative results.
    \item \textit{Multiple Grapes} (Fig.~\ref{fig:scene2}): The performance comparison is very close in this ideal scenario. While the baseline achieves similar results, it remains slightly inferior.
    \item \textit{Single Grape} (Fig.~\ref{fig:scene3}): Our approach shows poorer performance initially due to the grape's location in a dense, unknown area, requiring an initial entropy exploration phase. The baseline may find the cluster by chance. However, our method ultimately achieves a better F1 Score by the end of the planning steps.
    \item \textit{Unoriented Start} (Fig.~\ref{fig:scene4}): Starting from a recovery position, our method performs nearly twice as well as the baseline. As with the \textit{Full Occlusion} benchmark, the semantic 3D occupancy map is notably richer, as reflected in the quantitative results.
\end{enumerate}
Fig.~\ref{fig:experimental_setup} shows a qualitative comparison in the \textit{Unoriented Start} scenario for real-world deployment. Starting from the same unoriented initial robot configuration, our approach effectively recovers a meaningful pose, outperforming the predefined approach. Additional qualitative results for real-world deployment are detailed in Table~\ref{tab:merged_quantitative_results} and are available in the multimedia documentation of the manuscript.

\vspace{-0.3em}
\section{Conclusion}
\vspace{-0.1em}
We presented a benchmark for Active Vision in agricultural robotics, focusing on NBV planning. Our system was tested in both simulation and real-world scenarios, using ZSL to ensure perception independence. Future work will focus on deploying the system on a mobile agricultural robot and improving 3D reconstruction, potentially using techniques like Gaussian Splatting or NeRF to optimize Structure From Motion initialization and viewpoints selection.

These advancements can enhance agriculture by improving resource management and productivity, critical for meeting rising food demands. Accurate fruit counting in occluded environments is key for large-scale farming facing labor shortages. This research enhances agricultural environment perception via Active Vision and Zero-Shot Learning, automating repetitive tasks, and promoting efficiency and sustainability through robotics deployment.




\bibliographystyle{IEEEtran}
\bibliography{IEEEabrv, mybib}

\end{document}